\documentclass[twocolumn,journal]{IEEEtranTIE}
\usepackage[T1]{fontenc}
\usepackage[latin9]{inputenc}
\usepackage{array}
\usepackage{refstyle}
\usepackage{float}
\usepackage{booktabs}
\usepackage{textcomp}
\usepackage{multirow}
\usepackage{graphicx}
\usepackage[unicode=true,
 bookmarks=true,bookmarksnumbered=true,bookmarksopen=true,bookmarksopenlevel=1,
 breaklinks=false,pdfborder={0 0 0},pdfborderstyle={},backref=false,colorlinks=false]
 {hyperref}
\hypersetup{pdftitle={Your Title},
 pdfauthor={Your Name},
 pdfpagelayout=OneColumn, pdfnewwindow=true, pdfstartview=XYZ, plainpages=false}

\makeatletter

\AtBeginDocument{\providecommand\Secref[1]{\ref{Sec:#1}}}
\AtBeginDocument{\providecommand\Figref[1]{\ref{Fig:#1}}}
\AtBeginDocument{\providecommand\Tabref[1]{\ref{Tab:#1}}}
\providecommand{\tabularnewline}{\\}
\RS@ifundefined{subsecref}
  {\newref{subsec}{name = \RSsectxt}}
  {}
\RS@ifundefined{thmref}
  {\def\RSthmtxt{theorem~}\newref{thm}{name = \RSthmtxt}}
  {}
\RS@ifundefined{lemref}
  {\def\RSlemtxt{lemma~}\newref{lem}{name = \RSlemtxt}}
  {}

\usepackage[caption=false,font=footnotesize]{subfig}
\usepackage{graphicx}
\usepackage{cite}
\usepackage{picinpar}
\usepackage{amsmath}
\usepackage{stfloats}
\usepackage{url}
\usepackage{flushend}
\usepackage{colortbl}
\usepackage{soul}
\usepackage{multirow}
\usepackage{pifont}
\usepackage{color}
\usepackage{alltt}
\usepackage{enumerate}
\usepackage{siunitx}
\usepackage{breakurl}
\usepackage{epstopdf}
\usepackage{pbox}

\makeatother

\begin{document}
\title{Depth Evaluation for Metal Surface Defects by Eddy Current Testing
using Deep Residual Convolutional Neural Networks}
\author{Tian~Meng,~Yang~Tao{*},~Ziqi~Chen,~Jorge~R.~Salas~Avila,~Qiaoye~Ran,~Yuchun~Shao,~Ruochen~Huang,~Yuedong~Xie,~Qian~Zhao,~
Zhijie~Zhang,~Hujun~Yin,~Anthony~ J.~Peyton,~and~Wuliang~Yin\thanks{Tian Meng, Yang Tao, Ziqi Chen, Jorge R. Salas Avila, Qiaoye Ran,
Yuchun Shao, Ruochen Huang, Hujun Yin, Antony J. Peyton, and Wuliang
Yin are with the Department of Electrical and Electronic Engineering,
School of Engineering, The University of Manchester, Manchester, M13
9PL, U.K. Yuedong Xie is with the School of Instrumentation and Optoelectronic
Engineering, Beihang University, Beijing, 100191, China. Qian Zhao
is with the College of Engineering, Qufu Normal University, Shandong,
273165, China. Zhijie Zhang is with the School of Instrument and Electronics,
North University of China, Taiyuan, Shanxi, 030051, China. The corresponding
author is Yang Tao (mchikyt3@gmail.com).}}
\markboth{{\tiny{}This work has been submitted to the IEEE for possible publication.
Copyright may be transferred without notice, after which this version
may no longer be accessible.}}{}
\maketitle
\begin{abstract}
Eddy current testing (ECT) is an effective technique in the evaluation
of the depth of metal surface defects. However, in practice, the evaluation
primarily relies on the experience of an operator and is often carried
out by manual inspection. In this paper, we address the challenges
of automatic depth evaluation of metal surface defects by virtual
of state-of-the-art deep learning (DL) techniques. The main contributions
are three-fold. Firstly, a highly-integrated portable ECT device is
developed, which takes advantage of an advanced field programmable
gate array (Zynq-7020 system on chip) and provides fast data acquisition
and in-phase/quadrature demodulation. Secondly, a dataset, termed
as MDDECT, is constructed using the ECT device by human operators
and made openly available. It contains 48,000 scans from 18 defects
of different depths and lift-offs. Thirdly, the depth evaluation problem
is formulated as a time series classification problem, and various
state-of-the-art 1-d residual convolutional neural networks are trained
and evaluated on the MDDECT dataset. A 38-layer 1-d ResNeXt achieves
an accuracy of 93.58\% in discriminating the surface defects in a
stainless steel sheet. The depths of the defects vary from 0.3 mm
to 2.0 mm in a resolution of 0.1 mm. In addition, results show that
the trained ResNeXt1D-38 model is immune to lift-off signals.
\end{abstract}


\section{Introduction}

\IEEEPARstart{E}{ddy} current testing (ECT) is a non-destructive
testing (NDT) method harnessing the principle of electromagnetic induction,
which, compared to other NDT methods, has the virtue of high speed,
low cost and no contact \cite{sophian2001electromagnetic}. These
features make ECT an attractive technique in the detection and evaluation
of surface defects for conductive materials \cite{garcia-martin2011nondestructive}.
Recovering the profiles of a defect, e.g. location and depth, from
eddy current (EC) signals is a major topic in the research of ECT,
where machine learning (ML) plays an important role \cite{binali2017reviewon}.

Conventional ML algorithms have been adopted in various ECT applications,
and many of these studies generally utilised a two-step approach.
Firstly, raw EC signals would be subject to a feature transform or
extraction process, such as Principle Component Analysis \cite{kim2010classification},
time-frequency analysis by Rihaczek Distribution \cite{hosseini2012application},
Wavelet Transform \cite{smid2005automated}, Hilbert\textendash Huang
Transform \cite{liu2017studyon}, geometry recognition from Lissajous
Figure \cite{yin2019anovel}, and Convolutional Sparse Coding \cite{tao2019defectfeature}.
Next, the resultant feature representations, in order to achieve the
ultimate task of detecting and classifying defects, would be fed to
a classification or clustering algorithm such as Support Vector Machine
(SVM) \cite{smid2005automated,liu2017studyon,yin2019anovel}, Multi-Layer
Perceptron \cite{smid2005automated}, K-Means \cite{kim2010classification,hosseini2012application},
K-Nearest Neighbours \cite{smid2005automated,yin2019anovel}, Decision
Tree \cite{yin2019anovel} and Naive Bayes \cite{yin2019anovel}.
While these conventional ML algorithms still remain vibrant today
in the research of ECT, Deep Learning (DL) methods prevail more recently,
encouraged by their remarkable success in many other areas such as
image classification.

A Deep Belief Network was exploited in \cite{bao2020adeep} so as
to, from the EC scan images of the defects on the surface of a Titanium
sheet, extract features that were then fed to a least-square SVM algorithm
to classify the defects. The dataset was also evaluated in \cite{deng2020defectimage}
with a plain Convolutional Neural Network (CNN), which, in contrast
to \cite{bao2020adeep} where the feature extractor and classifier
were separate, was trained end-to-end. In \cite{li2019learning},
an encoder-decoder CNN, named EddyNet, was proposed aiming at learning
an inverse model, which predicted a crack profile given an EC signal.
Training samples were procured from a forward model with inputs and
outputs exchanged. In terms of pulsed ECT, a multi-task CNN was developed
in \cite{fu2019towards}, which installed a softmax layer and a fully
connection layer as two outputs in order to simultaneously classify
the type and predict the depth of flaw, respectively. In \cite{demachi2020crackdepth},
a plain CNN was used to estimate the crack depth for a heat transfer
tube of the steam generator of a pressurised water reactor, which,
compared to conventional numerical models, was less computationally
expensive at inference time. These DL-motivated studies all entailed
a larger dataset compared to those using conventional ML algorithms.
Specifically, the numbers of training samples in \cite{fu2019towards,demachi2020crackdepth,li2019learning}
were more than twenty thousand, while those in \cite{hosseini2012application,kim2010classification,liu2017studyon,smid2005automated,tao2019defectfeature,yin2019anovel}
were mostly a few hundreds. In \cite{deng2020defectimage,li2019learning,fu2019towards,demachi2020crackdepth},
CNN was used which was one of the most popular networks in DL research.
Nonetheless, the adopted CNNs were wide and shallow, which was at
variance with the `deep' feature of modern neural networks.

The recent advancement of CNN was largely driven by the ImageNet Large-Scale
Visual Recognition Challenge (ILSVRC) \cite{deng2009imagenet}. AlexNet
\cite{krizhevsky2012imagenet}, the winner in 2012, was regarded as
a break-through and drew attention on CNNs. In 2014, two very deep
CNNs emerged. The first one was GoogLeNet (with Inception modules)
\cite{szegedy2015goingdeeper}, which adopted a sparsely connected
architecture of stacking Inception modules composed of filters of
various sizes. In contrast, the second, VGG \cite{simonyan2015verydeep},
exploited smaller filters of the same size for all the convolutional
layers and increased the depth. Both very deep CNNs were able to achieve
compelling performances, however, they usually suffered the degradation
problem \cite{srivastava2015highway} that training accuracy would
saturate and then degrade as the depth increased. In 2015, ResNet
\cite{he2016deepresidual} was proposed to address the degradation
problem and won the ILSVRC-2015 with an ultra-deep network of 152
layers. The fundamental idea was to let the network fit a residual
mapping, instead of the original propagation, by adding skipping connections
between some layers, so that in principle a deep network would not
produce higher training error than its shallower counterpart. The
additional shortcut connections enabled gradients to propagate backwards
to earlier layers more easily, and hence resulted in easier training
than VGG. Later, ResNet evolved to the second version \cite{he2016identity},
where in each unit the activation layer preceded the convolutional
layer. In 2016, ResNeXt \cite{xie2017aggregated} was proposed, which,
in the residual module, harnessed the split-transform-merge pattern
akin to the Inception module. These revolutionary CNN architectures
have influenced many deep networks in applications beyond image classification.
In particular, the state-of-the-art residual CNNs such as ResNet and
ResNeXt can be applied to the research of defect depth estimation
with ECT; however, it has not been seen in the literature.

In this paper, the problem of estimating the depth of a surface defect
of a metallic sheet is addressed using a new ECT device and the state-of-the-art
DL techniques. The main contributions are three-fold. Firstly, a portable
multi-functional ECT device is introduced, which integrates an field
programmable gate array (FPGA), an ARM processor and the Windows 10
operating system. Secondly, the defect depth estimation problem is
formulated as a time series classification problem, and a dataset
using the ECT device is constructed and made openly available. We
name the dataset as MDDECT (Metal Defects of different Depths by ECT)
and aim to initiate a data-sharing campaign. It can serve as a testbed
and would encourage advancing the research of ECT in light of modern
DL techniques. Lastly, the state-of-the-art residual CNNs are applied
for the first time, to our knowledge, in the research of ECT. An accuracy
rate of 93.58\% is achieved using a 38-layer 1-d ResNeXt for defects
with a depth resolution of 0.1 mm in a stainless steel sheet.

The remainder of the paper is organised as follows. \Secref{Residual-CNN-Architecture}
presents the architectures of the 1-d residual CNNs. The hardware
design of the integrated ECT device is described in \Secref{Portable-ECT-Device}.
\Secref{ECT-Dataset} introduces the procedures of data collection
and the details of the MDDECT dataset. The configurations of hyper-parameters
and training process are demonstrated, along with the results of different
CNNs and discussions in \Secref{DCNN-Training-Experiment}. The last
section concludes the work and suggests further research directions.

\section{Architecture of 1-D Residual CNN\label{sec:Residual-CNN-Architecture}}

A residual CNN, e.g. ResNet, is constructed by stacking `residual
units', which learns a residual function $\mathcal{R}(x):=\mathcal{F}(x)-x$
where $\mathcal{F}(x)$ is the original underlying mapping \cite{he2016deepresidual}.
Formally, a residual unit conducts calculations as expressed in (\ref{eq:res_block1}),
in which $x_{l}$ and $y_{l}$ are the input and output of the $l^{th}$
residual unit, respectively, and $\sigma$ is an activation function.
The block $\mathcal{R}$ takes $x_{l}$ as input, and performs transformations
with weights $W_{l}$. In the original version of ResNet, the activation
function $\sigma$ is a rectified linear unit (ReLU). The block $\mathcal{R}$
is chosen from either a stack of two convolution units or a `bottleneck'
unit. A convolution unit is a convolution layer followed by a batch
normalisation (BN) layer and a ReLU layer. A bottleneck unit comprises
three convolution units, with the first and last convolution layers
being 1$\times$1 convolutions, which are used to reduce dimensionality
hence computational complexity.

In the second version of ResNet, the residual unit performs calculations
as shown in (\ref{eq:res_block1-1}), where the activation is an identity
function \cite{he2016identity}. However, the convolution unit in
the block $\mathcal{R}$ is pre-activated, that is, the BN and ReLU
layers precede the convolution layer. It was verified in \cite{he2016identity}
that (\ref{eq:res_block1-1}) enabled gradients to propagate to any
layers more easily than (\ref{eq:res_block1}). In ResNeXt \cite{xie2017aggregated},
the residual unit performs calculations as shown in (\ref{eq:res_block1-1-1}),
where the block $\mathcal{R}$ is an aggregation of a number of transformations
$\mathcal{T}_{i}$, and the number $C$ is the cardinality. In practice,
the split-transform-merge block $\mathcal{R}$ is usually implemented
using an equivalent grouped convolution, where the number of groups
equals the cardinality $C$. It is noted that here we assume that
ResNeXt inherits from the second version of ResNet. In addition, the
input $x_{l}$ and output $y_{l}$ share the same dimension in (\ref{eq:res_block1}),
(\ref{eq:res_block1-1}) and (\ref{eq:res_block1-1-1}) so as to illustrate
the idea. When the dimensions are different, a convolution layer will
replace the identity connection to match the dimensions.

\begin{equation}
y_{l}=\sigma\left(x_{l}+\mathcal{R}(x_{l},W_{l})\right)\label{eq:res_block1}
\end{equation}

\begin{equation}
y_{l}=x_{l}+\mathcal{R}(x_{l},W_{l})\label{eq:res_block1-1}
\end{equation}

\begin{equation}
y_{l}=x_{l}+\sum_{i=1}^{C}\mathcal{T}_{i}(x_{l},W_{li})\label{eq:res_block1-1-1}
\end{equation}

\begin{figure}[H]
\begin{centering}
\includegraphics[width=1\columnwidth]{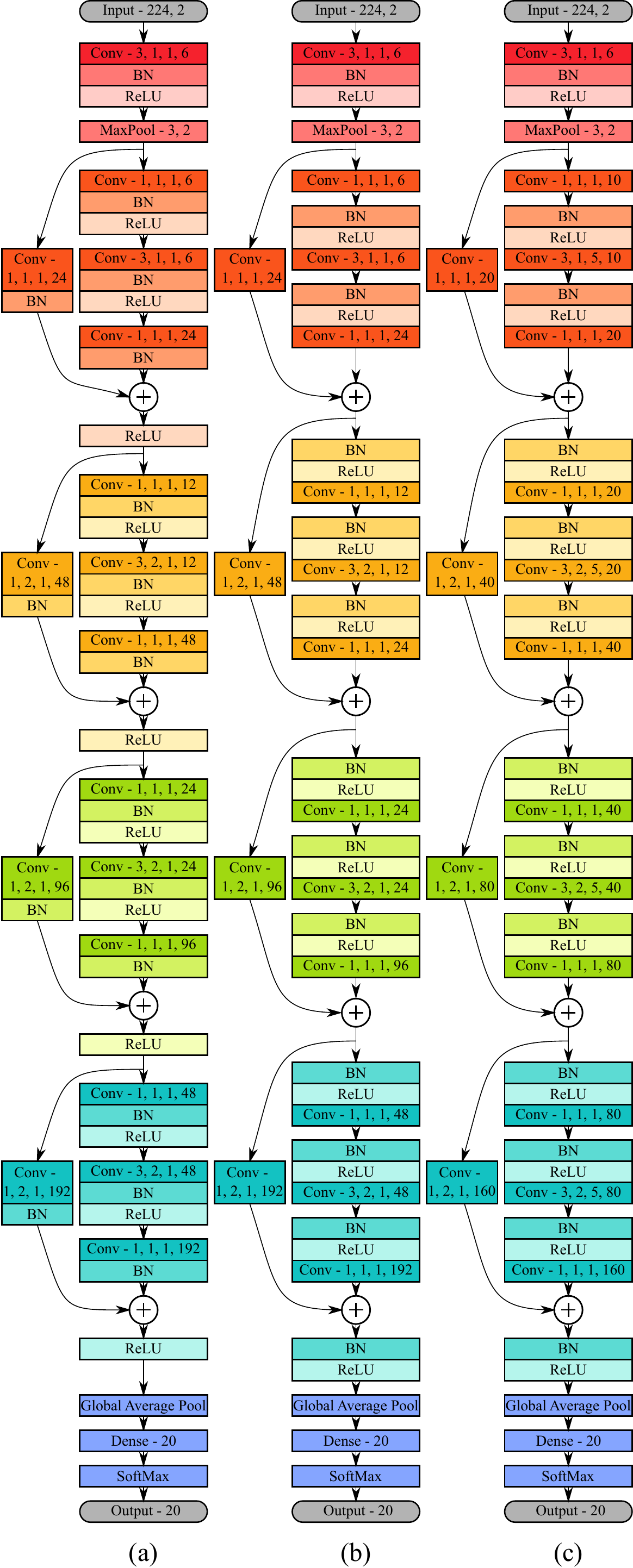}
\par\end{centering}
\caption{Detailed architectures of three 1-d residual networks. Four stages
in each network are marked in different colours. The input tensor
has 224 sampling points and 2 channels. In terms of the convolution
layer, the numbers in the block represent the kernel size, strides,
number of groups and number of filters. In terms of the max pooling
layer, the numbers represent the kernel size and strides. The output
tensor has 20 labels. (a), (b) and (c) correspond to ResNet1Dv1-14,
ResNet1Dv2-14 and ResNeXt1D-14, respectively. \label{fig:ResNet-Block-1}}
\end{figure}

The convention of naming a network is followed in this paper by appending
the version and depth to the type of network. However, because the
convolution layers used are 1-d instead of 2-d, we append `1D' to
the name in order to differentiate the networks from the original
2-d ones. Usually, residual CNNs comprise multiple stages, each one
of which has one or a stack of multiple residual units. In this paper,
we unify the number of stages to four. The first residual unit of
each stage doubles the channel dimension while halves the temporal
dimension. In order to clarify details, \Figref{ResNet-Block-1} illustrates
the architectures of ResNet1Dv1-14, ResNet1Dv2-14 and ResNeXt1D-14,
which serve as the base-line networks for the defect depth classification
task. They all have the same depth of 14 essential layers.

\section{Hardware Design of ECT Device\label{sec:Portable-ECT-Device}}

\begin{figure*}[tbh]
\begin{centering}
\includegraphics[width=2\columnwidth]{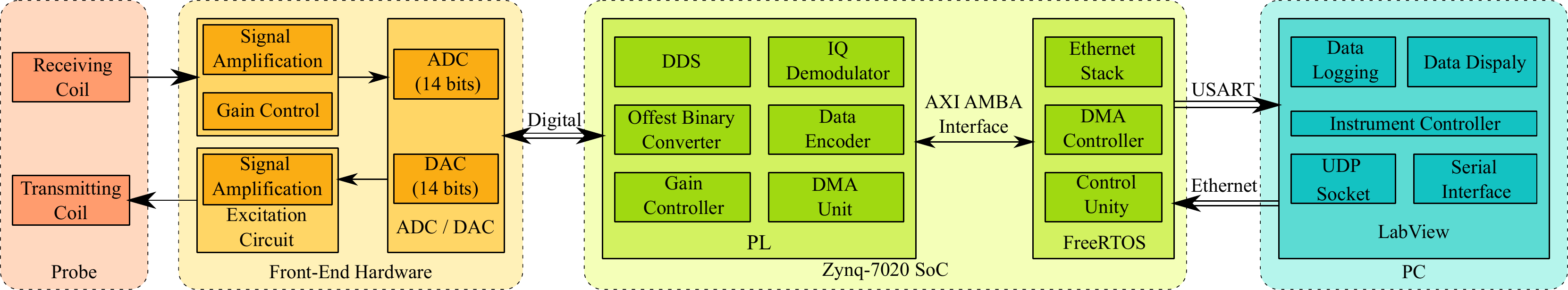}
\par\end{centering}
\caption{The block diagram of the architecture of the ECT device. The system
is mainly composed of four modules, including a replaceable coil probe
sensor, an SoC composed of an FPGA and an ARM processor, front-end
circuits and a host PC running Windows 10 system. \label{fig:Architecture-of-the}}
\end{figure*}

\begin{figure*}[tbh]
\begin{centering}
\includegraphics[width=2\columnwidth]{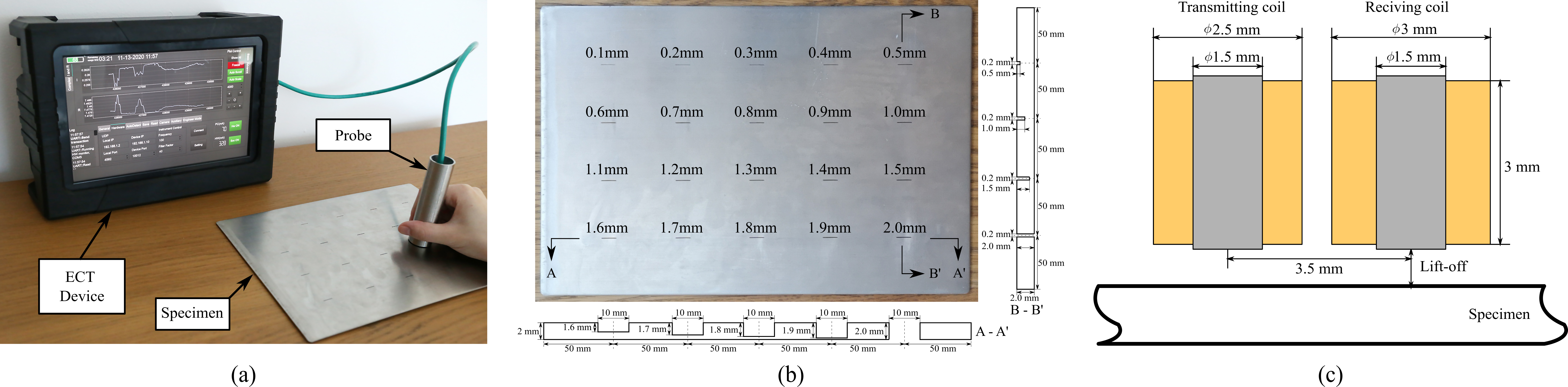}
\par\end{centering}
\caption{Images of the ECT device, specimen and probe sensor schematic. (a)
illustrates a scene of a human operator scanning a steel sheet with
a hand-hold probe sensor. (b) demonstrates the geometry dimensions
of a steel sheet that is used as a specimen to collect data from.
(c) shows the schematic of a probe sensor, which consists of two cylindrical
ferrite-cored coils.\label{fig:Images-of-the}}
\end{figure*}

The architecture of the ECT device is shown in \Figref{Architecture-of-the}.
The system mainly consists of four components, which are a replaceable
coil probe sensor, a Zynq-7020 system on chip (SoC), front-end circuits
and a host PC. Zynq-7020 SoC is the cornerstone of the system, which
integrates an ARM dual Cortex-A9 processor and a Xilinx 7-series FPGA.
This module is responsible for generating excitation signals, implementing
in-phase and quadrature (I/Q) demodulation and transferring data between
the module and the host PC. The front-end circuits consist of ADC/DAC,
signal amplification and gain control modules. A parallel digital
interface is exploited to connect the front-end hardware and the SoC
via an FPGA Mezzanine Card (FMC) connector. The system is capable
of providing a multi-frequency excitation signal, and the received
signal can be demodulated at each frequency simultaneously. The signal-to-noise
ratio (SNR) of the output signal is above 80 dB on average. \Figref{Images-of-the}(a)
demonstrates a scene of a human operator holding a probe sensor and
scanning a stainless steel sheet. The received signals and statistic
information are displayed on the screen of the ECT device in real-time.

\section{MDDECT Dataset\label{sec:ECT-Dataset}}

The success of deep learning in an application relies heavily on the
effort of constructing large and well-labeled datasets. For instance,
ImageNet contains 14 million high-quality images in 22 thousand visual
categories \cite{deng2009imagenet}. However, it is difficult in principle
to develop a universal dataset like ImageNet for defect depth estimation
by ECT, because the impedance signal captured by an ECT sensor is
determined by a plurality of factors. A widely accepted standard specimen
with standard defects is lacking in the ECT research community.

The MDDECT dataset was constructed using the integrated ECT device
described in \Secref{Portable-ECT-Device}, which installed a probe
sensor as shown in \Figref{Images-of-the}(c). The excitation frequency
was set as 20 kHz taking into account the system SNR and skin depth
of the plate under test. The signal data rate was configured as 2,500
samples per second. A stainless steel sheet with 20 machine-fabricated
slots on the surface was used as the specimen to scan, whose detailed
geometry dimensions are shown in \Figref{Images-of-the}(b). The defects
were surface opening cracks, and shared the same length and width
of 10 mm and 0.2 mm, respectively. The depth of the defects started
from 0.1 mm and incremented by 0.1 mm to the largest 2.0 mm. The defect
of 2.0 mm in depth was a through crack, as the thickness of the sheet
was 2.0 mm.

\begin{figure*}[tbh]
\begin{centering}
\includegraphics[width=2\columnwidth]{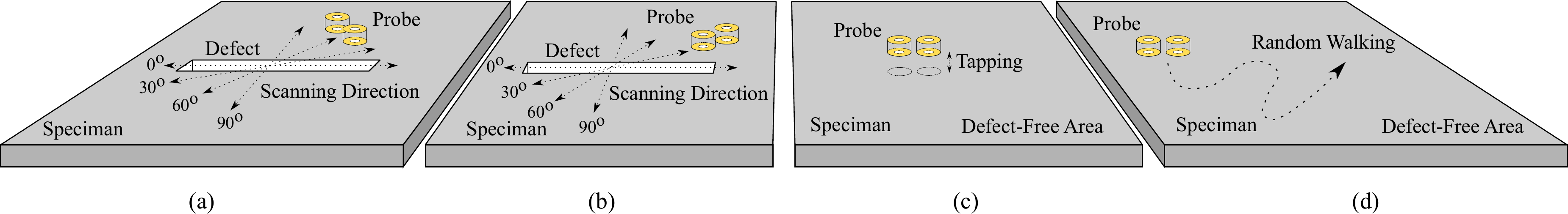}
\par\end{centering}
\caption{Illustrations of data collecting processes. (a) and (b) demonstrate
the probe scanning across a defect in eight different angles and two
directions (back and forth). Notice that the line crossing the axial
centres of the two coils is either tangental or normal to the scanning
direction. (c) shows the probe tapping on the surface of a defect-free
area in order to generate lift-off signals. (d) shows the probe moving
in a random trajectory on the surface of a defect-free area in order
to generate normal signals.\label{fig:Procedure}}
\end{figure*}

Although the MDDECT dataset was specialised to the ECT device and
the specimen, many other practical variances, which would nontrivially
affect the performance of an ECT system, were taken into account.
Firstly, thirty volunteers, who had no experience operating an ECT
device, were invited to scan the defects, hence introducing a great
variety of uncertainties, as composed to some research, e.g. \cite{yin2019anovel,tao2019defectfeature,bao2020adeep,deng2020defectimage},
where an automatically controlled movement was harnessed to scan defects.
Secondly, lift-off signals were deliberately collected and labeled,
so that the classifier should be able to differentiate lift-offs and
defects. As seen in \Figref{Procedure}(c), lift-off signals were
generated by randomly tapping the probe to a defect-free area on the
surface of the specimen. The lift-off distances were controlled to
be under 3 mm above the surface. In addition, normal signals of defect-free
areas were also captured and labeled, and the capturing process is
illustrated in \Figref{Procedure}(d). Thirdly, eight different scanning
angles between the long-edge line of a defect and the line crossing
the axial centres of the two coils were determined, as demonstrated
in \Figref[s]{Procedure}(a) and \ref{fig:Procedure}(b). Also, the
volunteers scanned across a defect along an angle in two directions,
back and forth. Lastly, when scanning a defect, the volunteers were
asked to try to maintain a constant lift-off distance of 0.5 mm, constant
moving speed of 60 mm/s and hold the probe vertical to the surface.
However, variances were inevitable and represented more practical
testing scenarios, hence making the MDDECT challenging.

After a preliminary test, it was found that the signals of the defects
of 0.1 mm and 0.2 mm were under the noise floor, and hence the data
from these two defects were excluded from the dataset. As a result,
the total number of classes was 20, including 18 classes of defects,
the lift-off class and normal signal class. Each volunteer repeated
the same scanning 5 times. All scans were divided into scan segments,
each of time window 0.5 seconds, which gave rise to the temporal dimension
of 1250. Ultimately, the dataset tensor had dimensions of (30, 8,
2, 5, 20, 1250, 2), each one of which represented the number of volunteers,
scanning angles, directions, repeats of each scanning, classes, temporal
points, and channels, respectively. The last dimension corresponded
to the in-phase and quadrature channels. In total, the MDDECT comprehended
48,000 scan segments (or scans for simplicity) in 20 classes. In terms
of the split of training and test sets, volunteers and the data scanned
by them were randomly selected. As a result, the dimensions for training
and test sets are (43200, 1250, 2) and (4800, 1250, 2), respectively\footnote{The numbers 43,200 and 4,800 are calculated from 27$\times$8$\times$2$\times$5$\times$20
and 3$\times$8$\times$2$\times$5$\times$20, respectively.}, constituting 90\% and 10\% of the total samples. Hence the test
set consists of three randomly selected volunteers. The MDDECT dataset
is available on Kaggle\footnote{https://www.kaggle.com/mchikyt3/mddect}.

\section{Experiment Details and Results\label{sec:DCNN-Training-Experiment}}

Before training a network, we needed to extract and set aside a validation
set from the training set, in order to determine hyper-parameters
and select DL models. The test set must not be used in the training
phase, and should only serve to produce the final claim on accuracy.
As the training set contains data from 27 volunteers, we randomly
selected 3 from the 27 volunteers and used their corresponding data
as a validation set. In addition, the temporal dimension was decimated
from 1250 to 250 by a factor of 5, in order to enable a faster training.
As a result, the dimensions of the training, validation and test sets
were (38400, 250, 2), (4800, 250, 2) and (4800, 250, 2), respectively.
The ratio of the number of samples among them was 8:1:1.

\subsection{Normalise and augment data}

\begin{figure}[tbh]
\begin{centering}
\includegraphics[width=0.95\columnwidth]{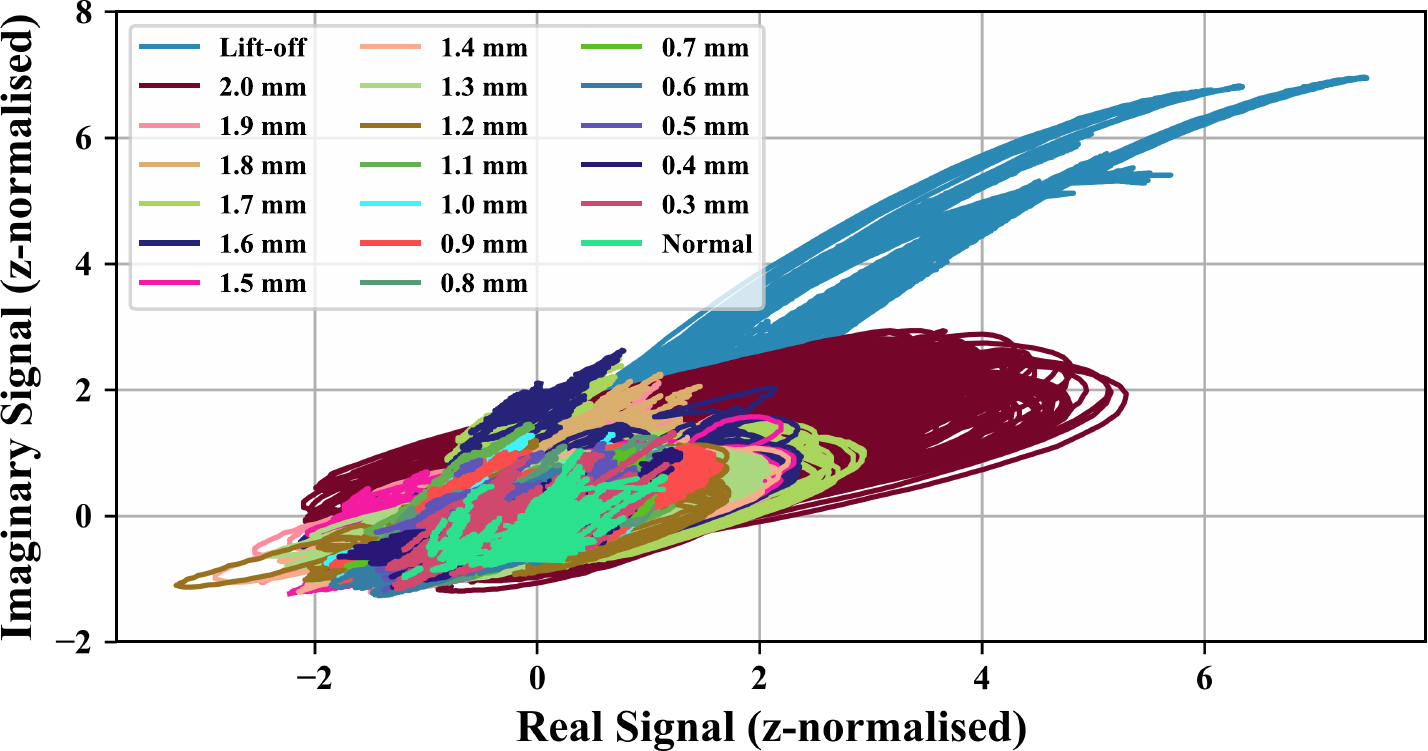}
\par\end{centering}
\caption{The z-normalised validation data all together in a complex plane.
The mean and standard deviation used in the normalisation were calculated
from the training data. The x and y axes represent the in-phase and
quadrature channels, respectively.\label{fig:Signal-Comparison}}
\end{figure}

The final training data was applied with a channel-wise normalisation
according to (\ref{eq:z-norm}), where $x_{i}$ is the flattened tensor
when the channel dimension equals $i$, and $\mu_{i}$ and $\sigma_{i}$
are the mean and standard deviation of $x_{i}$, respectively. This
normalisation is termed as z-normalisation \cite{ismailfawaz2019deeplearning},
which nullifies the mean and standardises the variance of the data
in terms of each channel. The resultant training data was then used
to train a network. In addition, the calculated $\mu_{i}$ and $\sigma_{i}$
from the training data were used to normalise the validation and test
data too. In order to appreciate the data in general before training,
the z-normalised validation data are plotted all together in a complex
plane in \Figref{Signal-Comparison}, from which it can be seen that
the lift-off signals are larger in magnitude and also different in
phase compared to the defect signals. However, most signals overlap
severely with each other, indicating the difficulties to classify
these signals.

\begin{equation}
\frac{x_{i}-\mu_{i}}{\sigma_{i}},i=1,2\label{eq:z-norm}
\end{equation}

After normalisation, the data was then configured for train-time and
test-time augmentations. In terms of train-time augmentation, every
training sample was cropped randomly for every epoch in order to introduce
certain variances to the training data on the fly. Concretely, a segment
of dimensions (224, 2) was cropped randomly under a uniform distribution
from the original training sample of dimensions (250, 2). As a result,
the input dimensions to a network were (224, 2), where the batch dimension
was not shown. In terms of test-time augmentation, 10-crop test was
conducted to the validation and test samples, that is, the final classification
result of a sample was determined by averaging the outputs of a network
for 10 random crops from the sample.

\begin{table*}[!t]
\caption{Architecture Details of Residual CNNs\label{tab:deeperandwider}}
\resizebox{2.\columnwidth}{!}{

\begin{tabular}{cccccccccc}
\toprule 
\multirow{3}{*}{Stage} & \multirow{3}{*}{Output Size} & \multicolumn{8}{c}{Network Architecture}\tabularnewline
\cmidrule{3-10} \cmidrule{4-10} \cmidrule{5-10} \cmidrule{6-10} \cmidrule{7-10} \cmidrule{8-10} \cmidrule{9-10} \cmidrule{10-10} 
 &  & \multirow{2}{*}{ResNet1Dv1-26} & \multirow{2}{*}{ResNet1Dv2-26} & \multirow{2}{*}{ResNeXt1D-26} & ResNet1Dv1-14 & ResNet1Dv2-14 & \multirow{2}{*}{ResNeXt1D-14 (Wider 1)} & \multirow{2}{*}{ResNeXt1D-14 (Wider 2)} & \multirow{2}{*}{ResNeXt1D-38}\tabularnewline
 &  &  &  &  & (Wider) & (Wider) &  &  & \tabularnewline
\midrule 
 & 224 & conv - 3, 1, 1, 6 & conv - 3, 1, 1, 6 & conv - 3, 1, 1, 6 & conv - 3, 1, 1, 8 & conv - 3, 1, 1, 8 & conv - 3, 1, 1, 8 & conv - 3, 1, 1, 8 & conv - 3, 1, 1, 6\tabularnewline
\cmidrule{2-10} \cmidrule{3-10} \cmidrule{4-10} \cmidrule{5-10} \cmidrule{6-10} \cmidrule{7-10} \cmidrule{8-10} \cmidrule{9-10} \cmidrule{10-10} 
 & 112 & \multicolumn{8}{c}{max pool - 3, 2}\tabularnewline
\midrule
1 & 112 & $\left[\begin{array}{c}
1,6\\
3,6\\
1,24
\end{array}\right]\times2$ & $\left[\begin{array}{c}
1,6\\
3,6\\
1,24
\end{array}\right]\times2$ & $\left[\begin{array}{c}
1,10\\
3,10,C=5\\
1,20
\end{array}\right]\times2$ & $\left[\begin{array}{c}
1,8\\
3,8\\
1,32
\end{array}\right]\times1$ & $\left[\begin{array}{c}
1,8\\
3,8\\
1,32
\end{array}\right]\times1$ & $\left[\begin{array}{c}
1,14\\
3,14,C=7\\
1,28
\end{array}\right]\times1$ & $\left[\begin{array}{c}
1,15\\
3,15,C=5\\
1,30
\end{array}\right]\times1$ & $\left[\begin{array}{c}
1,10\\
3,10,C=5\\
1,20
\end{array}\right]\times3$\tabularnewline
\midrule
2 & 56 & $\left[\begin{array}{c}
1,12\\
3,12\\
1,48
\end{array}\right]\times2$ & $\left[\begin{array}{c}
1,12\\
3,12\\
1,48
\end{array}\right]\times2$ & $\left[\begin{array}{c}
1,20\\
3,20,C=5\\
1,40
\end{array}\right]\times2$ & $\left[\begin{array}{c}
1,16\\
3,16\\
1,64
\end{array}\right]\times1$ & $\left[\begin{array}{c}
1,16\\
3,16\\
1,64
\end{array}\right]\times1$ & $\left[\begin{array}{c}
1,28\\
3,28,C=7\\
1,56
\end{array}\right]\times1$ & $\left[\begin{array}{c}
1,30\\
3,30,C=5\\
1,60
\end{array}\right]\times1$ & $\left[\begin{array}{c}
1,20\\
3,20,C=5\\
1,40
\end{array}\right]\times3$\tabularnewline
\midrule
3 & 28 & $\left[\begin{array}{c}
1,24\\
3,24\\
1,96
\end{array}\right]\times2$ & $\left[\begin{array}{c}
1,24\\
3,24\\
1,96
\end{array}\right]\times2$ & $\left[\begin{array}{c}
1,40\\
3,40,C=5\\
1,80
\end{array}\right]\times2$ & $\left[\begin{array}{c}
1,32\\
3,32\\
1,128
\end{array}\right]\times1$ & $\left[\begin{array}{c}
1,32\\
3,32\\
1,128
\end{array}\right]\times1$ & $\left[\begin{array}{c}
1,80\\
3,80,C=7\\
1,160
\end{array}\right]\times1$ & $\left[\begin{array}{c}
1,60\\
3,60,C=5\\
1,120
\end{array}\right]\times1$ & $\left[\begin{array}{c}
1,40\\
3,40,C=5\\
1,80
\end{array}\right]\times3$\tabularnewline
\midrule
4 & 14 & $\left[\begin{array}{c}
1,48\\
3,48\\
1,192
\end{array}\right]\times2$ & $\left[\begin{array}{c}
1,48\\
3,48\\
1,192
\end{array}\right]\times2$ & $\left[\begin{array}{c}
1,80\\
3,80,C=5\\
1,160
\end{array}\right]\times2$ & $\left[\begin{array}{c}
1,64\\
3,64\\
1,256
\end{array}\right]\times1$ & $\left[\begin{array}{c}
1,64\\
3,64\\
1,256
\end{array}\right]\times1$ & $\left[\begin{array}{c}
1,160\\
3,160,C=7\\
1,320
\end{array}\right]\times1$ & $\left[\begin{array}{c}
1,120\\
3,120,C=5\\
1,240
\end{array}\right]\times1$ & $\left[\begin{array}{c}
1,80\\
3,80,C=5\\
1,160
\end{array}\right]\times3$\tabularnewline
\midrule
 & 1 & \multicolumn{8}{c}{global average pool, 20-d fc, softmax}\tabularnewline
\midrule
\multicolumn{2}{c}{\# Trainable Parameters} & $9.37\times10^{4}$ & $9.30\times10^{4}$ & $9.38\times10^{4}$ & $1.01\times10^{5}$ & $1.00\times10^{5}$ & $9.77\times10^{4}$ & $1.14\times10^{5}$ & $1.35\times10^{6}$\tabularnewline
\multicolumn{2}{c}{FLOPs} & $3.70\times10^{6}$ & $3.69\times10^{6}$ & $3.84\times10^{6}$ & $4.11\times10^{6}$ & $4.09\times10^{6}$ & $4.25\times10^{6}$ & $4.99\times10^{6}$ & $5.42\times10^{6}$\tabularnewline
\bottomrule
\end{tabular}

}
\end{table*}

\subsection{Determine network architectures and hyper-parameters}

As discussed in \Secref{Residual-CNN-Architecture}, the networks
exploited in this paper were one-dimensional variants of the residual
CNNs. In addition, the CNNs were fixed to have four stages, and hence
the minimal depth was 14 for each network, where only one residual
module existed in each stage. \Figref{ResNet-Block-1} illustrates
the architectures of ResNet1Dv1-14, ResNet1Dv2-14 and ResNeXt1D-14,
which served as the base-line networks. The width of them, that is,
the number of filters in the first convolution layer, was set to 6.
They all had a similar level of trainable parameters and floating
point operations (FLOPs).

Based upon the three base-line networks, we experimented on the depth
dimension and doubled the number of residual modules in each stage
of the networks, which gave rise to three deeper networks: ResNet1Dv1-26,
ResNet1Dv2-26 and ResNeXt1D-26. In addition, we also attempted to
expand the width dimension from the base-line networks while maintaining
the number of trainable parameters and FLOPs similar to the 26-layer
ones, which gave rise to four wider networks. The details of the architectures
of these seven new networks are listed in \Tabref{deeperandwider}.
As they had a similar computation complexity, we would be able to
compare their performances and see whether depth or width was more
effective under the scope of this paper. Lastly, we evaluated our
deepest network ResNeXt1D-38, where in each stage there were three
residual modules. The details of the architecture of ResNeXt1D-38
are listed in the last column in \Tabref{deeperandwider}.

\begin{figure}[tbh]
\begin{centering}
\includegraphics[width=1\columnwidth]{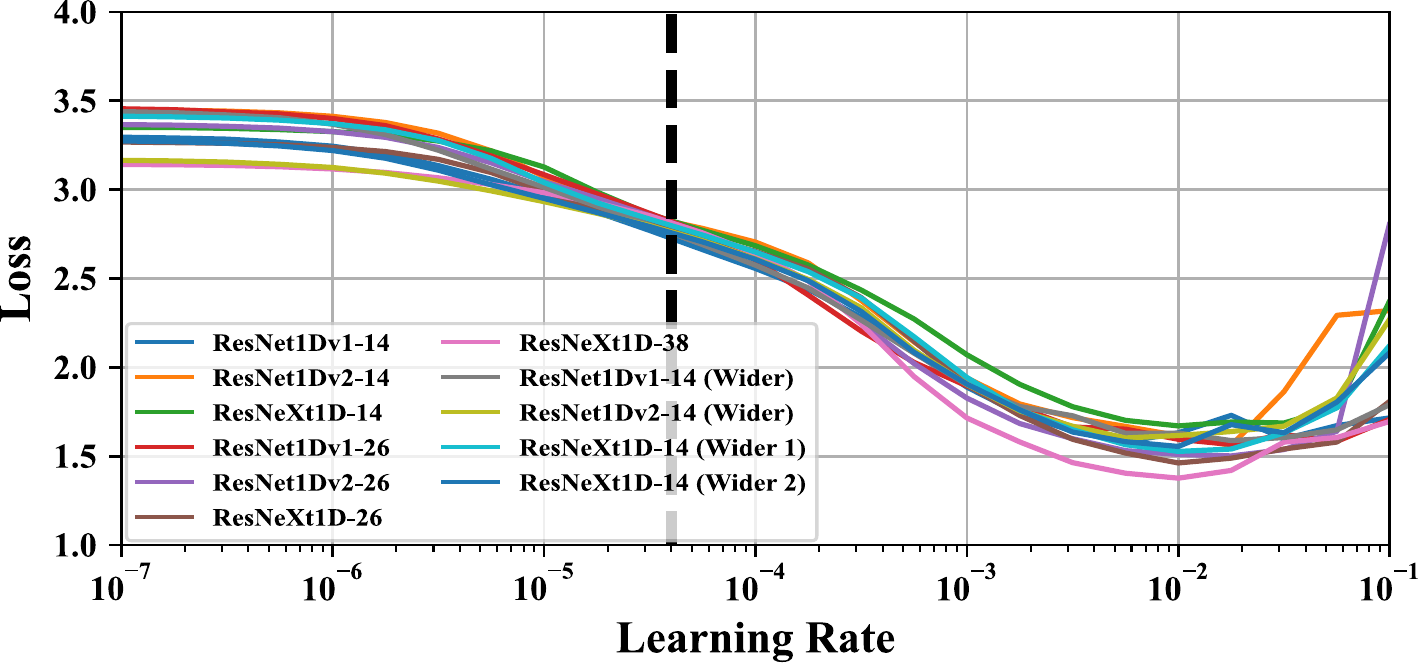}
\par\end{centering}
\caption{Losses v.s. learning rate. The best initial learning rate appears
to be the same for all the evaluated networks, which is about 4.0$\times$10$^{-5}$
and marked by a dotted vertical line in the figure.\label{fig:Data-Augmentation-2}}
\end{figure}

Learning rate is an important hyper-parameter affecting the training
and performance of a network. In order to find the best initial learning
rate, we performed the strategy from \cite{smith2017cyclical}, where
the training started with a small learning rate and increased it epoch
by epoch in a geometric progression. After a few epochs, the loss
v.s. learning rate plot could be drawn, and the best initial learning
rate located at the point where the loss decreased most rapidly. This
strategy was applied to all the networks we evaluated, and the resultant
plots of loss v.s. learning rate are shown in \Figref{Data-Augmentation-2}.
It can be seen that all the networks appear to share the same best
learning rate 4.0$\times$10$^{-5}$, at which the losses descend
rapidly. It is noted that this way of finding the best initial learning
rate may not an exact solution. Notice that the x-axis in the figure
is in a logarithmic scale.

All the networks were applied with the same following training configurations
and hyper-parameters. Adam optimiser was exploited with default values
of parameters recommended in \cite{kingma2015adama}, and the mini-batch
size was set to 128. The training data were shuffled for every epoch,
and each network was trained for 10,000 epochs. The learning rate
was initialised to 4.0$\times$10$^{-5}$, and decreased to 4.0$\times$10$^{-6}$
at epoch 5,000, and finally to 4.0$\times$10$^{-7}$ at epoch 7,500.

\begin{figure}[tbh]
\begin{centering}
\includegraphics[width=1\columnwidth]{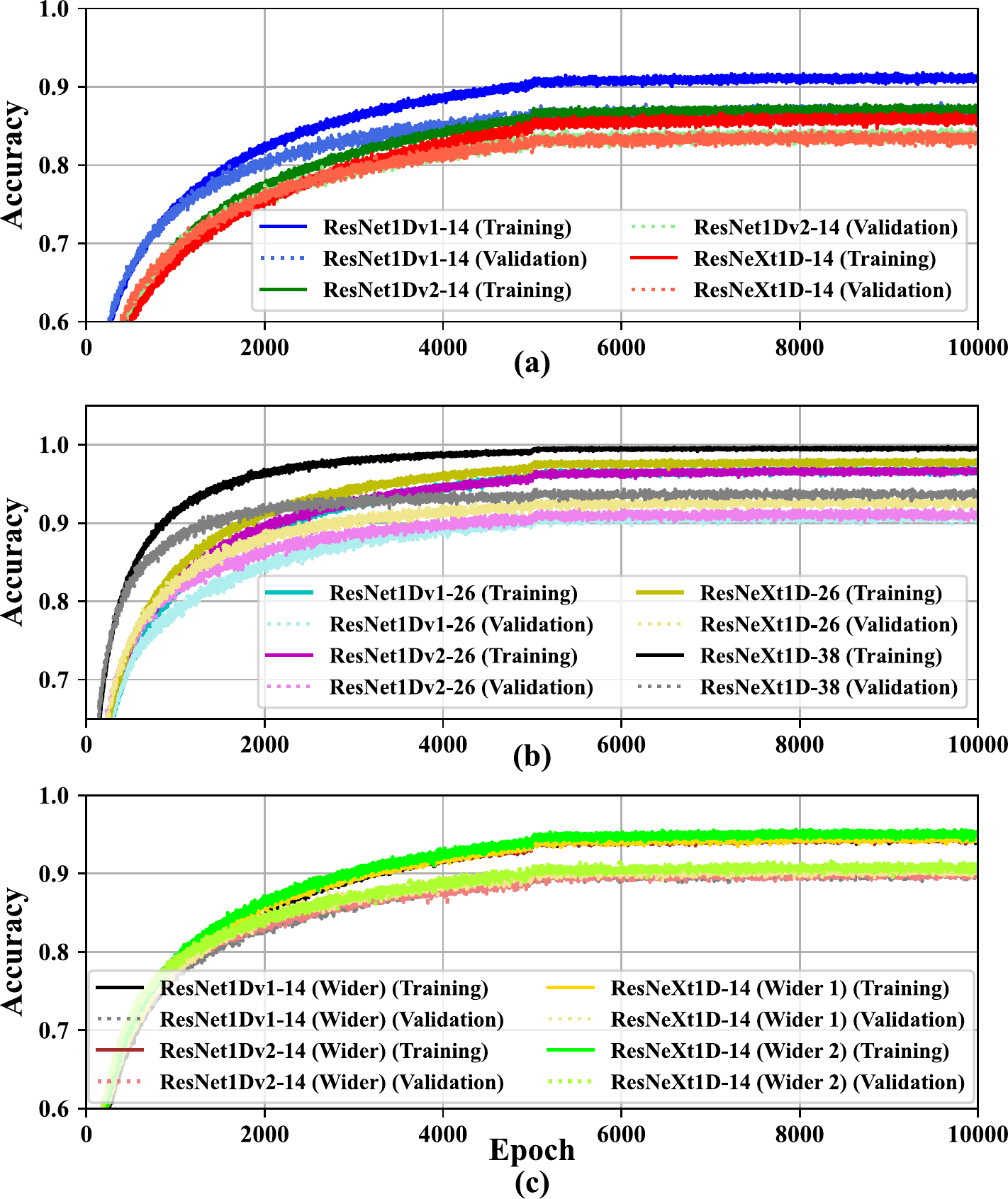}
\par\end{centering}
\caption{Training processes of eleven networks. (a), (b) and (c) depict the
training and validation accuracies for three base-line networks, four
deeper networks and four wider networks, respectively. Solid and dashed
lines represent the training and validation accuracies, respectively.
Results of the same network use similar colours. \label{fig:Training-on-MD}}
\end{figure}

\subsection{Train networks and analyse results}

The training was coded using the Tensorflow library, and conducted
on an Nvidia RTX 2080Ti GPU. In total, the training of the networks
took about five days to finish. The training and validation accuracies
and losses are displayed in \Figref{Training-on-MD} for three base-line
networks: ResNet1Dv1-14, ResNet1Dv2-14 and ResNeXt1D-14, four deeper
networks: ResNet1Dv1-26, ResNet1Dv2-26, ResNeXt1D-26 and ResNeXt1D-38,
and four wider ResNet1Dv1-14, ResNet1Dv2-14 and ResNeXt1D-14 with
respect to training epoch, which are also available as a TensorBoard
experiment\footnote{https://tensorboard.dev/experiment/hNURDaEzRr2GQCZeDyqBHw/}.

Three main observations can be taken from the training processes.
Firstly, the three base-line networks all suffer the under-fitting
problem as per \Figref{Training-on-MD}(a), that is, they converge
to large training errors between 15\% and 10\%, which implies that
their representation powers are insufficient for the MDDECT dataset.
Such suggestion is verified by the results of larger networks in \Figref{Training-on-MD}(b)
and \ref{fig:Training-on-MD}(c), where the training errors of ResNet1Dv1-26,
ResNet1Dv2-26 and ResNeXt1D-26 are well below 5\%, and the training
errors of the wider ResNet1Dv1-14, ResNet1Dv2-14 and ResNeXt1D-14
are slightly higher than 5\%. On the other hand, the validation errors
of the larger networks are around 10\%. However, it's difficult to
conclude that the larger networks are over-fitted to the training
data, only based on the about 5\% difference between the training
and validation errors. Secondly, the deeper networks in general have
achieved higher accuracies than the wider ones that share similar
level of trainable parameters and FLOPs, as can be seen by comparing
\Figref{Training-on-MD}(b) and \ref{fig:Training-on-MD}(c). Thirdly,
different versions and types of network lead to marginal differences
in terms of both the training and validation accuracies. In \Figref{Training-on-MD}(b),
the discrepancies of the training and validation accuracies of ResNet1Dv1-26,
ResNet1Dv2-26 and ResNeXt1D-26 are within 3\%. The discrepancies in
\Figref{Training-on-MD}(c) are even smaller. However, ResNeXt1D-26
has achieved the best performance among the larger networks of similar
levels of complexity. As a consequence, we further trained a ResNeXt1D-38,
the deepest network we evaluated, so as to test the limit of performance.
Its training and validation accuracies are plotted in \Figref{Training-on-MD}(b).

\begin{table}[tbh]
\begin{centering}
\caption{Classification Results\label{tab:Results}}
\par\end{centering}
\centering{}%
\begin{tabular}{ccc}
\toprule 
\multirow{2}{*}{Network Architecture} & \multicolumn{2}{c}{Accuracy}\tabularnewline
\cmidrule{2-3} \cmidrule{3-3} 
 & Top-1 & \textpm{} 0.1mm\tabularnewline
\midrule 
ResNet1Dv1-14 & 87.88\% & 95.14\%\tabularnewline
ResNet1Dv2-14 & 83.21\% & 93.01\%\tabularnewline
ResNeXt1D-14 & 86.65\% & 94.51\%\tabularnewline
ResNet1Dv1-26 & 92.50\% & 96.90\%\tabularnewline
ResNet1Dv2-26 & 91.85\% & 96.06\%\tabularnewline
ResNeXt1D-26 & 93.15\% & 96.78\%\tabularnewline
ResNet1Dv1-14 (Wider) & 90.42\% & 95.88\%\tabularnewline
ResNet1Dv2-14 (Wider) & 89.93\% & 95.65\%\tabularnewline
ResNeXt1D-14 (Wider 1) & 91.31\% & 96.23\%\tabularnewline
ResNeXt1D-14 (Wider 2) & 89.38\% & 95.00\%\tabularnewline
\textbf{ResNeXt1D-38} & \textbf{93.58\%} & \textbf{97.20\%}\tabularnewline
\bottomrule
\end{tabular}
\end{table}

\begin{figure}[tbh]
\begin{centering}
\includegraphics[width=0.9\columnwidth]{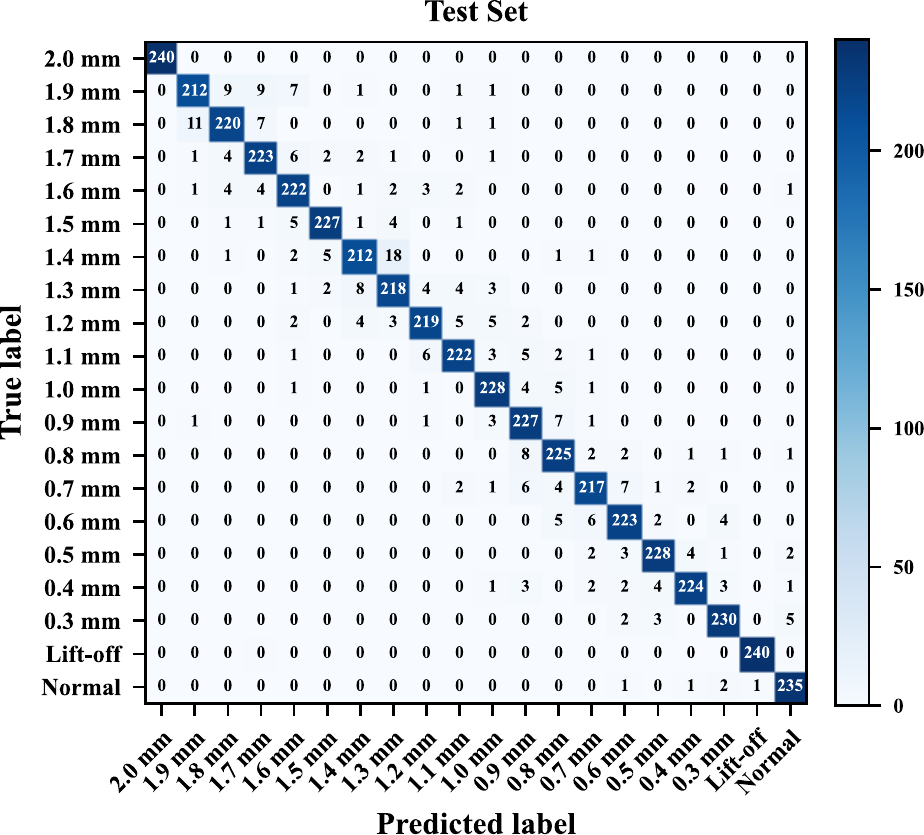}
\par\end{centering}
\caption{Confusion matrix of the trained ResNeXt1D-38 on the test set. Every
entry in the matrix represents the number of samples that are classified
to a specific class. \label{fig:Confusion-Matrix-2-1}}
\end{figure}

\begin{figure}[tbh]
\begin{centering}
\includegraphics[width=0.95\columnwidth]{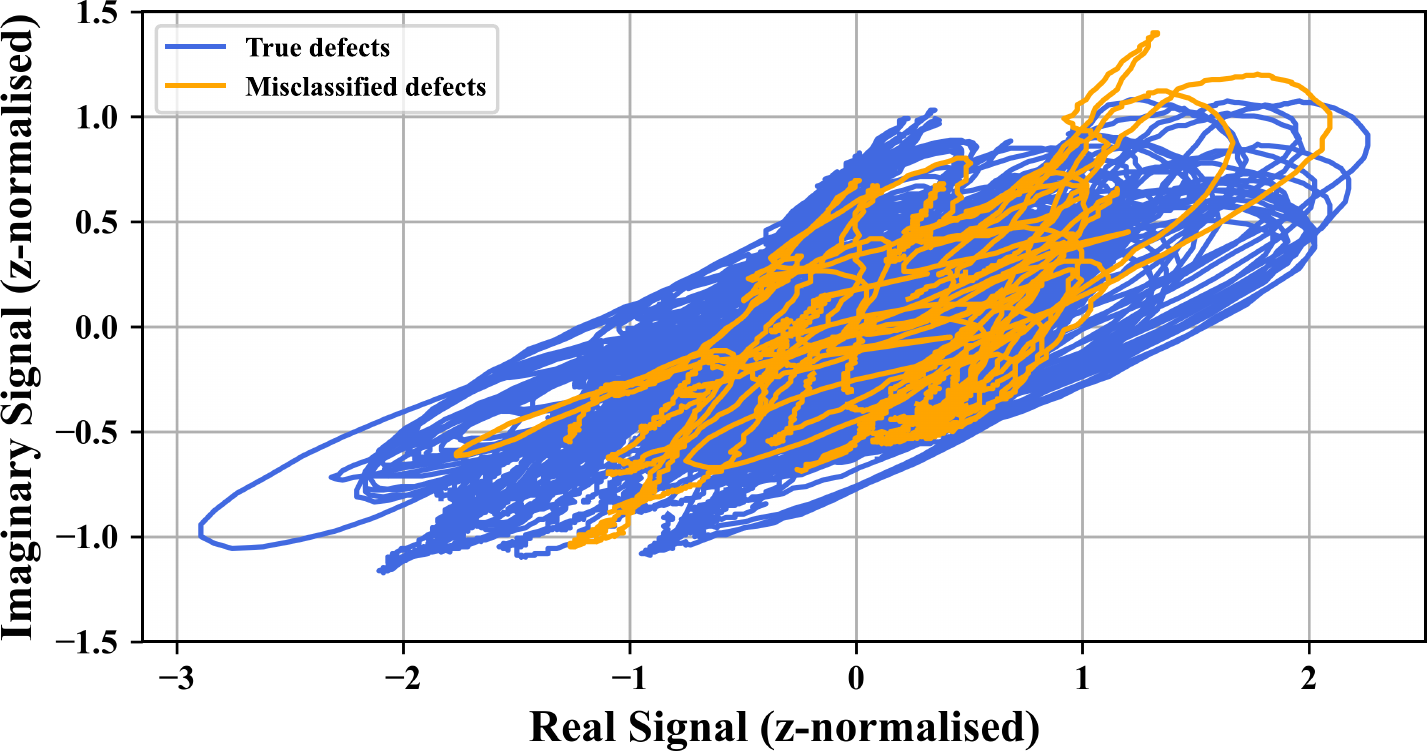}
\par\end{centering}
\caption{The z-normalised signals in the complex plane for the samples in the
test set that are predicted as the 1.4 mm defect from the trained
ResNeXt1D-38 model. Blue and yellow traces represent the true and
misclassified samples, respectively.\label{fig:Plots-of-the}}
\end{figure}

\begin{figure}[tbh]
\begin{centering}
\includegraphics[width=0.93\columnwidth]{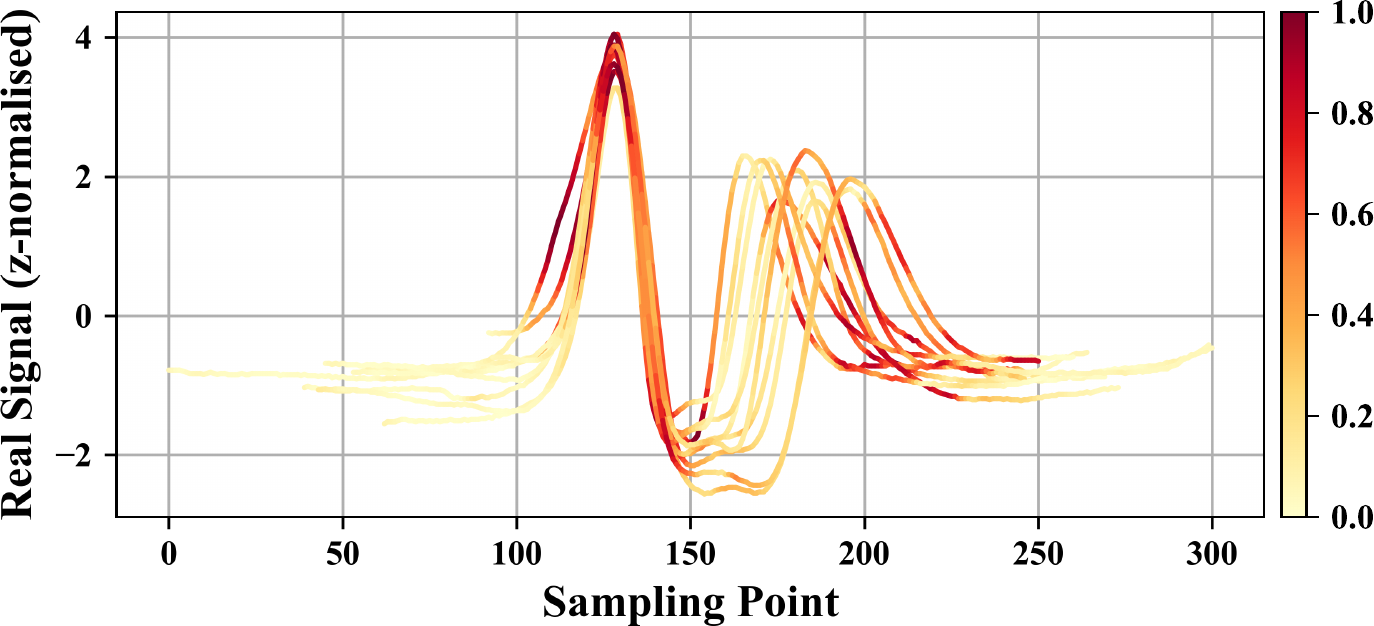}
\par\end{centering}
\caption{The real signals of the samples of the 2.0 mm defects in the test
set with class activation mappings calculated based on the trained
ResNeXt1D-38. The colour represents the activation level of a sampling
point with respect to the 2.0 mm defect class. The plots are aligned
according to their first peaks in order to clarify the activation
regions.\label{fig:Class-Activation-Mapping}}
\end{figure}

In order to claim the final accuracies, the best trained model of
a network was selected as the one that achieved the highest validation
accuracy during training for that network. Next, the test set data
was fed to the chosen model to produce the final claimed accuracy
of the network. The final top-1 accuracies of each network are listed
in \Tabref{Results}. It can be seen that ResNeXt1D-38 has achieved
the highest accuracy of 93.58\%, while the second best accuracy is
93.15\%, achieved by ResNeXt1D-26. The fact that the additional 12
layers give rise to an improvement of only 0.43\% implies that simply
increasing the depth may not be able to push the boundary of performance,
and over-fitting may occur. Moreover, if the accuracy metric is relaxed
to tolerate an error of $\pm$ 0.1 mm, the `$\pm$ 0.1 mm accuracies'
are also listed in the same table, where ResNeXt1D-38 also wins with
97.20\%. If we examine closer on the results of ResNeXt1D-38, its
confusion matrices on the test set is shown in \Figref{Confusion-Matrix-2-1}.
It can be seen that the mistaken samples tend to be classified into
the adjacent classes of their ground-truth classes, which explains
the much improved 97.20\% accuracy with $\pm$ 0.1 mm tolerances.
In addition, it appears that shallower defects are not more difficult
to classify than deeper ones. Lastly, the lift-off samples are all
correctly detected. Being immune to lift-off signals is an important
and desirable feature for an ECT defect depth estimation method.

In order to further understand the misclassified samples, we can plot
the data according to their predicted labels. For instance, the test
set data labeled as the 1.4 mm defect by the trained ResNeXt1D-38
are plotted in \Figref{Plots-of-the}, from which it can be seen that,
the wrongly labeled samples overlap in great deal with the correct
samples. This phenomenon can be found for other defects as well. We
argue that it is almost impossible to estimate by human the depth
of a defect in the resolution of 0.1 mm, while ResNeXt1D-38 has achieved
a 93.58\% accuracy. A Class Activation Mapping (CAM) can be calculated
as per \cite{zhou2016learning} to examine the contributions of every
sampling points to the classification of a time series. As an illustration,
CAMs for the samples of the 2.0 mm defect in the test set were calculated
based on the trained ResNeXt1D-38, and are shown in \Figref{Class-Activation-Mapping}.
It can be seen that only some regions of the time series are activated
and other regions are masked out. This can help explain why ResNeXt1D-38
is able to correctly differentiate samples of different classes that
have overlapping parts, because they have different activation regions
for their own classes. Moreover, it is worth pointing out that the
times series of the same class share similar activation regions, as
can be clearly seen in \Figref{Class-Activation-Mapping}.

\section{Conclusions}

The evaluation of the depth of a surface defect of metallic materials
is a major application of ECT, where, recent DL-motivated methods
commence to surpass the conventional ones. However, many existing
approaches have not taken full advantage of the state-of-the-art DL
techniques proposed in computer vision. In this paper, we aim at addressing
the problem of ECT-based surface defect depth estimation by using
1-d deep residual convolutional networks. Firstly, a highly integrated
and multi-functional portable ECT device is developed based on Zynq-7020
SoC, which provides fast data acquisition and I/Q demodulation. Secondly,
a dataset, termed as the MDDECT, is constructed by 30 volunteer operators
using the ECT device, and consists of 48,000 samples of 20 classes
in total. The MDDECT dataset is openly available and can be exploited
as a testbed in order to promote new ECT algorithms. Thirdly, eleven
1-d residual networks of three different types are evaluated, and
a 38-layer network network ResNeXt1D-38 has achieved an accuracy of
93.58\% in terms of estimating the depth of the surface defects from
0.3 mm to 2.0 mm with depth resolution of 0.1 mm. In addition, the
deep learning algorithms can reject lift-off signal, hence immune
to lift-off noise. Future research directions would be to evaluate
a different family of deep networks such as recurrent networks and
other learning strategies. Moreover, a continued effort should be
made to enrich the MDDECT dataset with more diversities, scans and
specimens. 

\bibliographystyle{IEEEtranTIE}
\bibliography{reference}

\end{document}